\documentclass[lettersize,journal]{IEEEtran}
\usepackage{amsmath,amsfonts}
\usepackage{algorithm}
\usepackage{array}
\usepackage[colorlinks,
            linkcolor=blue,
            anchorcolor=blue,
            citecolor=blue,
            ]{hyperref}
\usepackage[caption=false,font=normalsize,labelfont=sf,textfont=sf]{subfig}
\usepackage{textcomp}
\usepackage{stfloats}
\usepackage{url}
\usepackage{verbatim}
\usepackage{graphicx}
\usepackage{cite}
\usepackage{bm}

\usepackage[capitalise,nameinlink]{cleveref}

\usepackage{algpseudocode}

\usepackage{multirow}
\usepackage{booktabs}

\usepackage{xcolor}
\usepackage{soul}

\begin{document}

\title{Learning Feature Encoder with Synthetic Anomalies for Weakly Supervised Graph Anomaly Detection}

\author{
Yingjie Zhou,~\IEEEmembership{Member,~IEEE,}
Yuqin Xie, 
Fanxing Liu,
Dongjin Song,~\IEEEmembership{Member,~IEEE,}
Ce Zhu,~\IEEEmembership{Fellow,~IEEE,}
Lingqiao Liu,~\IEEEmembership{Member,~IEEE}

\thanks{Yingjie Zhou, Yuqin Xie, and Fanxing Liu are with Sichuan University, Chengdu 610065, China (e-mail: yjzhou09@gmail.com; xieyuqin@stu.scu.edu.cn; liufanxing@stu.scu.edu.cn).}

\thanks{Dongjin Song is with the University of Connecticut, Storrs, CT 06269, USA (e-mail: dongjin.song@uconn.edu).}

\thanks{Ce Zhu is with the University of Electronic Science and Technology of China, Chengdu 611731, China (e-mail: eczhu@uestc.edu.cn).}

\thanks{Lingqiao Liu is with the University of Adelaide, Adelaide, SA 5005, Australia (e-mail: lingqiao.liu@adelaide.edu.au).}
\thanks{*Lingqiao Liu is the corresponding author.}
}


\markboth{the manuscript has been published by IEEE (IEEE TKDE, 2026, DOI: 10.1109/TKDE.2026.3656821). Please do not cite the preprint}%
{Zhou \MakeLowercase{\textit{et al.}}: Learning Feature Encoder with Synthetic Anomalies for Weakly Supervised Graph Anomaly Detection}


\maketitle

\begin{abstract}

Weakly supervised graph anomaly detection aims to unveil unusual graph instances, \textit{e.g.}, nodes, whose behaviors are significantly different from the normal ones, under the condition that only a limited number of annotated anomalies but abundant unlabeled samples are available.
A major challenge for this task is to learn a meaningful latent feature representation that reduces intra-class variance among normal data while remaining highly sensitive to anomalies. Although recent works have applied self-supervised feature learning methods for graph anomaly detection, their strategies are not specifically tailored to the unique requirements of graph anomaly detection, which motivates our exploration of a more domain-specific feature learning approach.
In this paper, we introduce a weakly supervised graph anomaly detection method that leverages a feature learning strategy specifically tailored for graph anomalies. Our approach is built upon a multi-task learning scheme designed to extract robust feature representations, through synthesized anomalies. We generate these synthetic anomalies by perturbing the normal graph in various ways and assign a dedicated detection head to each anomaly type. This design ensures that the learned features are sensitive to potential deviations from normal patterns.
Although synthetic anomalies may not perfectly replicate real-world patterns, they provide valuable auxiliary data for effective feature learning—much like the way features learned from classifying ImageNet images are used in various downstream computer vision tasks. Additionally, we adopt a two-phase learning strategy to balance the influence of synthetic anomalies and real data. The process begins with an initial warm-up phase using only synthetic samples, followed by a full-training phase that integrates both tasks.
Numerous experiments on public datasets demonstrate the superior performance of our proposed strategy, in comparison with those of its competitors.
Code is available at \url{https://github.com/yj-zhou/SAWGAD}.  

\end{abstract}


\section{Introduction}\label{sec:intro} 

\IEEEPARstart{G}{raph} anomaly detection has attracted increasing attention, due to its broad applications ranging from social network analysis \cite{yu2016survey, hu2013social, wang2019detecting} to financial fraud detection \cite{virdhagriswaran2006camouflaged, dou2020caregnn, shi2022h2, wang2019semignn} and intrusion monitoring \cite{jorjani2020graph, caville2022anomal}.
Constructing an effective detection model usually requires abundant labeled training samples that include both normal and abnormal instances.
However, the real-world scenarios typically encounter a severely imbalanced situation where only a limited number of annotated anomalies are available alongside abundant unlabeled samples \cite{ma2021comprehensive, gui2024survey}.
This phenomenon is inherent to the nature of anomaly detection, as anomalous behaviors occur infrequently in practice. More importantly, comprehensively identifying and labeling all potential anomalies is both resource-intensive and technically challenging, if not impossible\cite{jiang2023weakly, chandola2009anomaly}.

To address the challenge of limited labeled data, one promising approach is to leverage self-supervised feature learning methods to derive good feature representations without relying on labeled examples. As demonstrated across various deep learning applications, a well-learned feature representation can substantially reduce the need for vast amounts of training data\cite{jing2020self, yu2023self, gui2024survey, liu2022graph}. In the context of graph anomaly detection, we expect such features to both minimize intra-class variance among normal instances and remain highly sensitive to potential outliers. In line with this, several methods have adopted self-supervised approaches inspired by generic self-supervised learning algorithms—such as reconstruction-based objectives\cite{ding2019dominant, fan2020anomalydae, li2019specae, zhou2021subtractive, pei2022resgcn, ma2025salad} and contrastive learning strategies\cite{zheng2021slgad, liu2021cola, xu2022conad, chen2022gccad}—to tackle graph anomaly detection. Although previous efforts have made progress, their feature learning strategies are designed for generic graph representation, aiming to capture inherent patterns in data. Therefore, they are not specifically tailored to the graph anomaly detection task and do not fully meet the need for effective feature representations in this domain. This observation motivates us to investigate a more domain-specific approach to feature learning for graph anomaly detection.

\IEEEpubidadjcol

In this paper, we introduce a weakly supervised graph anomaly detection method that leverages a dedicated feature learning strategy tailored for graph data. Different from current methods that perform graph representation learning through self-supervised learning techniques or directly employing a GNN-based feature extractor, we conduct a multi-task learning as an auxiliary task for feature learning. The goal of the auxiliary task is to learn discriminative perturbation-sensitive features that help the model to distinguish between normal contexts and perturbed ones. Specifically, we create synthetic anomalies by perturbing the normal graph in various ways, then we design tasks that make the network detect each type of anomaly with a dedicated head. In this way, we enforce the features to be sensitive to the potential deviations from normal patterns. The key insight is that, although synthetic anomalies may not perfectly mimic real-world anomaly patterns, they serve as valuable auxiliary data for learning effective features—similar to how features learned from classifying ImageNet images are used for various downstream computer vision tasks. The role of synthetic anomalies is to make the feature encoder sensitive to various types of disturbances, enabling it to quickly adapt to detecting real anomalies through subsequent supervised fine-tuning. This is a feature learning strategy rather than an attempt to enumerate all anomaly types. Moreover, we employ a two-phase learning strategy to balance both synthetic anomalies and real data. This strategy involves an initial warm-up phase using only synthetic samples, followed by a full-training phase that integrates both tasks.

\begin{figure}[t]
\centering
\includegraphics[width=0.9\linewidth]{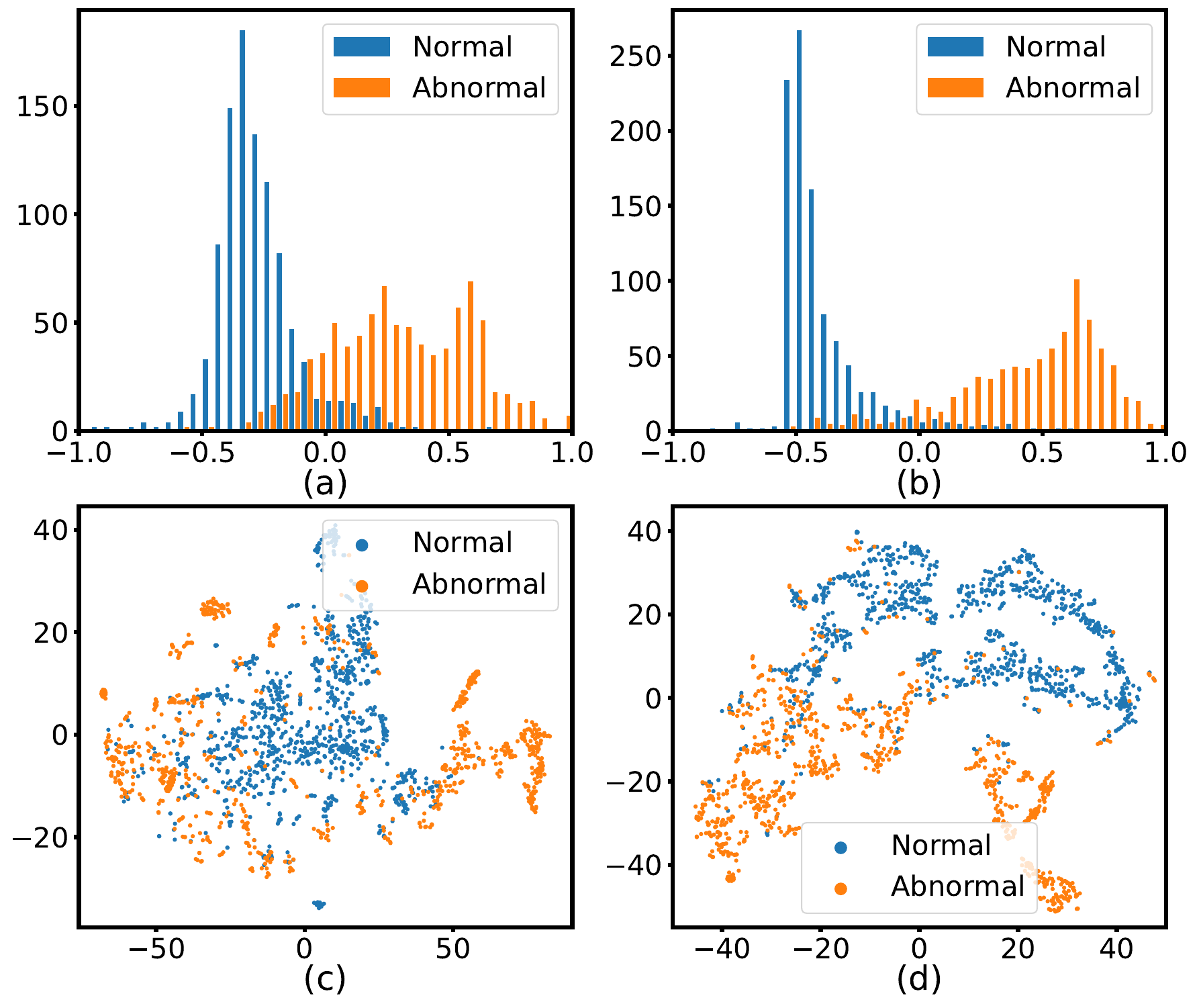}
\caption{Illustration of the visualization results without/with our proposed encoding strategy. The visualization is developed on the Weibo dataset. Figure(a)(b) demonstrate the histogram of the linear discriminant analysis (LDA) values, where the vertical axis represents the frequency (or count) of samples falling into each bin along the first LD component. Figure(c)(d) show the t-SNE visualization of the latent distributions. Both the LDA values and t-SNE visualization for the proposed encoding strategy (Figure(b)(d)) exhibit more distinguishable from those without it (Figure(a)(c)). The clear separation in the learned feature space demonstrates the effectiveness of our disturbance-sensitive feature learning approach in building discriminative representations that facilitate anomaly detection.}
\label{fig:visualize}
\end{figure}

As shown in \cref{fig:visualize}, the learned feature representations clearly discriminate the abnormal samples from the normal ones, enabling effective detection for graph anomalies.
Based on the learned latent representations, the anomaly
detector generates a probability for each input sample to determine whether it is anomalous. 
Experiments on various real-world datasets demonstrate that our anomaly detection model achieves significant improvements over existing competitive methods, particularly in scenarios with extremely limited number of labeled anomalies.

The main contributions of this paper are summarized as follows.

\begin{itemize}
    \item We introduce a novel feature
     encoding strategy that develops disturbance-sensitive feature representation through synthetic anomaly generation. To the best of our knowledge, this is the first in-depth study that explores the utilization of synthetic generated anomalies to guide latent feature learning for weakly supervised graph anomaly detection.
    
    \item We propose a simple yet effective graph anomaly detection model that employs specially designed multi-task learning to fulfill the proposed strategy, followed by the incorporated training of weakly supervised graph anomaly detection.

    \item Based on extensive experimental analysis on diverse public datasets, we demonstrate the superior performance of the proposed approach over its competitors. Ablation studies are also conducted to validate the contribution of each component in the proposed method.
\end{itemize}

The rest of this paper is structured as follows. Section~\ref{sec:related} reviews related works. In Section~\ref{sec:method}, we introduce our proposed model in details, including the synthetic anomaly generation process, feature encoding design, and the anomaly detection mechanism. To demonstrate the effectiveness of our method, extensive experiments are conducted in Section~\ref{sec:exp}. Finally, Section~\ref{sec:conclusion} concludes the paper.

\section{Related Works}\label{sec:related}

\subsection{Graph-based Anomaly Detection}
Graph anomaly detection (GAD) aims to identify nodes that exhibit significantly different patterns from the majority in terms of their structural connections and feature distributions\cite{ma2021comprehensive, qiao2024deep}. Traditional approaches\cite{akoglu2010oddball, li2014probabilistic, eswaran2018spotlight, pan2023prem} relied on handcrafted features and traditional machine learning methods. With the advent of graph neural networks (GNNs), deep learning based approaches have become the dominant paradigm due to their ability to automatically learn representations that capture both structural and feature information\cite{wu2020comprehensive}.

A typical problem in graph anomaly detection is camouflage \cite{virdhagriswaran2006camouflaged, hooi2017graph}, where fraudsters deliberately connect to normal nodes to disguise their abnormal patterns.
Methods like CARE-GNN\cite{dou2020caregnn} and PC-GNN\cite{liu2021pcgnn} propose selective neighbor aggregation strategies that carefully filter out suspicious neighbors during message passing. LGM-GNN\cite{li2023lgm} combines local and global memory networks to capture both neighbor-level and graph-level patterns for fraud detection. Some other works have explored the heterophilic nature of anomalous relations\cite{rogers1970homophily}, where anomalies often exhibit distinct connection patterns from normal nodes.
Several methods address heterophily through spectral approaches, such as BWGNN\cite{tang2022bwgnn} which leverages beta wavelet filters to capture high-frequency signals associated with anomalies, and GHRN\cite{gao2023ghrn} which establishes a connection between heterophily and graph frequency to prune inter-class edges. Other methods like H2-FDetector\cite{shi2022h2}, GATSep\cite{platonov2023gatsep}, and PMP\cite{zhuo2024partitioning} design specialized aggregation mechanisms that distinguish between homophilic and heterophilic connections, where PMP uses partitioned message passing to separately aggregate neighbors with different labels.
Recently, TAM\cite{qiao2024tam} exploits the observation that normal nodes show stronger affinity with each other than anomalous nodes do, while GDN\cite{gao2024gdn} tackles the structural distribution shift in heterophily between training and testing data by decomposing node features to better handle heterophilic connections. To better exploit the available normal nodes, GGAD\cite{qiao2024ggad} generates graph structure-aware outlier nodes as negative samples to train a discriminative one-class classifier.

While these methods have shown promising results, they typically assume access to abundant labeled data for both normal and anomalous nodes. However, this assumption rarely holds in real-world scenarios where obtaining labeled anomalies is extremely costly and time-consuming.

\subsection{Self-supervised Learning for Graph}

Self-supervised learning has emerged as a powerful paradigm for learning meaningful representations without explicit supervision\cite{gui2024survey}. This approach has been widely adopted in the graph domain, demonstrating impressive performance across various tasks\cite{liu2022graph}. Many successful approaches in graph self-supervised learning leverage reconstruction objectives. VGAE\cite{kipf2016variational} employs variational autoencoders to reconstruct graph structure, while GraphMAE\cite{hou2022graphmae} focuses on reconstructing masked node features through a carefully designed autoencoder. Another prominent line of research explores contrastive learning strategies. DGI\cite{velickovic2018dgi}, GMI\cite{peng2020gmi} and MVGRL\cite{hassani2020mvgrl} maximize the mutual information between different views of the graph, with the former two focusing on node features and topological structures, and the latter contrasting views generated through graph diffusion. GraphCL\cite{you2020graphcl} and GCC\cite{qiu2020gcc} further advance this direction by investigating various graph augmentation strategies and cross-network subgraph discrimination.

In the context of anomaly detection, self-supervised learning has also shown promising results by learning meaningful representations.
Methods like SpecAE\cite{li2019specae}, DOMINANT\cite{ding2019dominant}, ResGCN\cite{pei2022resgcn} and AnomalyDAE\cite{fan2020anomalydae} utilize reconstruction errors from graph autoencoders to identify anomalies, while SL-GAD\cite{zheng2021slgad}, CoLA\cite{liu2021cola}, CONAD\cite{xu2022conad}, GCCAD\cite{chen2022gccad} and HeCo\cite{wang2021heco} leverage contrastive learning to capture both structural and attribute deviations. However, these approaches are primarily unsupervised and do not utilize available label information about anomalies, which could provide valuable guidance for the detection task.

The most related approaches to our method are GraphCL\cite{you2020graphcl} and CONAD\cite{xu2022conad}. However, unlike methods using generic augmentation operations, our proposed method enables the model to learn perturbation-sensitive features that can quickly adapt to detecting real anomalies through subsequent supervised fine-tuning, even when real anomalies exhibit patterns different from synthetic ones.

\subsection{Weakly-supervised Anomaly Detection}

Weakly-supervised anomaly detection methods try to learn an anomaly detector with only a small number of labeled anomalies and a majority of the data unlabeled\cite{jiang2023weakly}. This setting is commonly seen in practice as anomalies are rare by nature and obtaining labeled examples often requires significant domain expertise and manual effort.

Several methods have been proposed in this field of research. Early methods focus on learning scoring functions. For example, DevNet\cite{pang2019devnet} directly optimizes anomaly scores by enforcing statistical deviation between normal and anomalous samples using a Gaussian prior, while Deep SAD\cite{ruff2020deepsad}, an extended version of Deep SVDD\cite{ruff2018deepsvdd}, learns a hypersphere boundary in the feature space. Building upon a novel feature encoding strategy, FEAWAD\cite{zhou2021feawad} introduces effective feature learning through autoencoders to enhance the detection performance. More recently, PReNet\cite{pang2023prenet} conducts an alternative approach by detecting anomalies through pairwise relation prediction between instances.

Recently, several graph anomaly detection approaches have also been developed specifically for the weakly-supervised setting. Most of them rely on pseudo-labeling or data augmentation techniques. ConsE\cite{chang2023conse} employs consistency-based neighbor sampling, utilizing attribute similarity and soft pseudo-labels.
BSL\cite{yu2024bsl} uses disentangled representation learning with consistency regularization where unlabeled nodes undergo various augmentations. To enhance the performance of graph anomaly detection, ConsisGAD\cite{chen2024consisgad} leverages learnable data augmentation and exploits homophily distribution differences between normal and anomalous nodes. CGNN\cite{li2025cgnn} addresses fraud detection with extremely limited labels through category semantic decomposition and denoising attention mechanism for adaptive neighbor aggregation. Data augmentation methods typically use perturbed samples to train the entire model, aiming to improve overall robustness and generalization. In contrast, our synthetic anomalies are designed specifically to guide feature learning in the encoder component only, making it sensitive to perturbations. Crucially, the synthetic anomaly data is not used to train the final anomaly detector—it serves solely as an auxiliary signal for learning discriminative feature representations.

To the best of our knowledge, the feature learning strategies in recent methods are not specifically tailored to graph anomaly detection under limited labeled data, thereby limiting the detection performance.

\section{THE PROPOSED APPROACH}\label{sec:method}

\subsection{Problem Definition}

Given an attributed graph $\mathcal{G} = (\mathcal{V}, \mathcal{E}, \mathbf{X})$, $\mathcal{V} = \{v_1, ..., v_n\}$ represents the set of nodes, $\mathcal{E}$
represents the set of edges, and $\mathbf{X} \in \mathbb{R}^{n \times d}$ represents node features where each node $v_i$ is associated with a $d$-dimensional feature vector $\mathbf{x}_i$. In weakly-supervised graph anomaly detection, we only observe binary labels $\mathbf{Y} = \{y_1, ..., y_m\}$ for a small subset of nodes $\mathcal{M} \subset \mathcal{V}$, where $m = |\mathcal{M}| \ll |\mathcal{V}|$, and $y_i \in \{0, 1\}$ with 1 indicating anomaly. The remaining nodes $\mathcal{U} = \mathcal{V} \setminus \mathcal{M}$ are unlabeled.

Our goal is to learn a function $f: \mathcal{V} \rightarrow [0,1]$ that estimates the probability of a node being anomalous, such that $f(v_i) > f(v_j), $ $\forall v_i, v_j \in \mathcal{V}: (v_i \in \mathcal{A}) \land (v_j \in \mathcal{N})$, where $\mathcal{A}$ denotes abnormal nodes and $\mathcal{N}$ denotes normal nodes.

\subsection{Overview}

\begin{figure*}[t]
\centering
\includegraphics[width=1\linewidth]{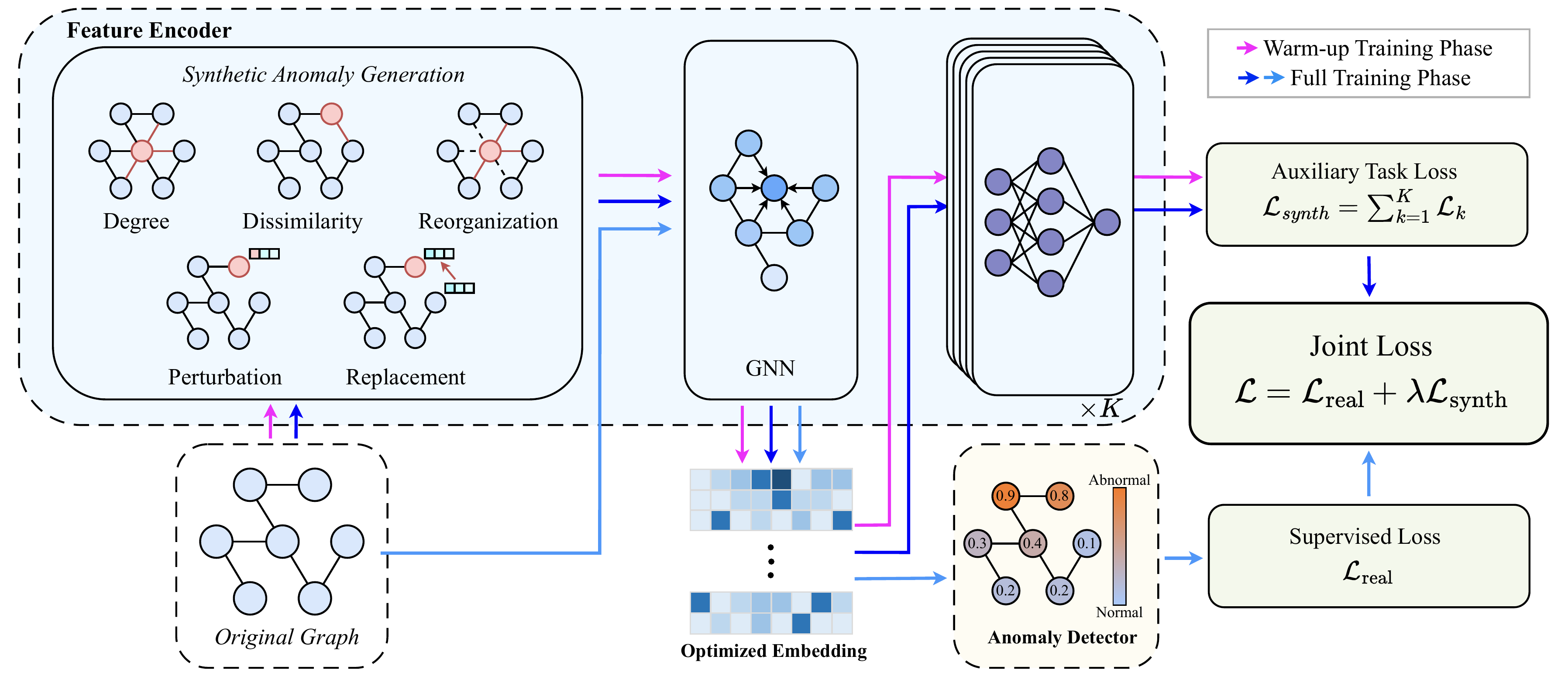}

\caption{Overview of the proposed approach. The model consists of two components, \textit{i.e.}, a feature encoder and an anomaly detector. 
The feature encoder employs specialized detection heads for diverse synthetic generated anomalies to develop disturbance-sensitive feature representations, which is further enhanced by real graph anomalies. The anomaly detector generates a probability for each input sample to indicate that whether it is an anomaly. In the fully trained phase, the two differently colored arrows denote parallel optimization paths, corresponding to two different optimization objectives in the loss function.}

\label{fig:overview}
\end{figure*}

We propose an approach for weakly supervised graph anomaly detection, leveraging synthetic anomalies to guide effective feature learning.
As illustrated in \cref{fig:overview}, our approach consists of two primary components, \textit{i.e.}, a feature encoder and an anomaly detector.
The feature encoder employs a multi-task learning process that utilizes diverse synthetic anomalies to develop disturbance-sensitive feature representations, helping the model become attuned to diverse types of abnormal patterns in graphs.
Real graph anomalies are also employed to further enhance the feature representation, while maintaining the model's sensitivity to various types of disturbances.
Riding on these refined representations, the anomaly detector generates a probability for each input sample to determine whether it is anomalous.
The training of the anomaly detector is jointly optimized with the feature encoder, in weakly supervised setting.
The following subsections provide detailed descriptions of each component.

\subsection{Feature Encoder}

The feature encoder contains a two-step process, \textit{i.e.}, synthetic anomaly generation, followed by a multi-task learning. The former step generates synthetic anomalies to enhance the latent representation learned in the latter step.

\subsubsection{Synthetic Anomaly Generation}

After thoroughly reviewing the literature on graph anomalies \cite{xu2022conad, bandyopadhyay2019outlier, song2007conditional, wang2019detecting}, we have summarized five types of representative abnormal behaviors that commonly occur in graph anomaly detection scenarios. We design corresponding perturbation operations that represent these types of anomalies by manipulating node features or graph structure. Each type targets a specific aspect of node characteristics, helping the model learn representations that are sensitive to different patterns of deviation observed in practice.

\paragraph{\textbf{Degree-based Structural Anomaly ($\tau_1$)}}
This type of anomaly exhibits unusual connectivity patterns \cite{xu2022conad}, where the degree of the node is much larger than those of the remaining nodes. In real-world scenarios, this may represent the behaviors of bot accounts in social networks that establish an abnormally large number of connections to spread misinformation, or the behaviors of fraudulent entities in financial networks creating numerous transactions to obscure money laundering activities.

Specifically, we generate additional edges connecting to the target node to enlarge its degree. The newly generated edges can be represented as:
\begin{equation}
\small
\Delta \mathcal{E}_v = \{(u, v) | u \sim \text{Uniform}(\mathcal{V}), |\Delta \mathcal{E}_v| = \alpha \cdot \sigma(\text{deg}(\mathcal{G}))\},
\end{equation}
where $\sigma(\text{deg}(\mathcal{G}))$ represents the standard deviation of node degrees in the graph, and $\alpha \sim \text{Uniform}(3,5)$ controls the intensity of the deviation.

\paragraph{\textbf{Dissimilarity-based Connection Anomaly ($\tau_2$)}}
This type of anomaly tends to connect highly dissimilar nodes\cite{song2007conditional, bandyopadhyay2019outlier}, thereby violating the homophily principle \cite{mcpherson2001birds} commonly observed in real networks. In practice, this could correspond to unusual interactions in academic citation networks where a paper cites works from completely unrelated research domains, potentially signaling academic misconduct.

Specifically, we calculate the pairwise feature distances $d(v,u)$ between the target node $v$ and nodes sampled from the candidate set $\mathcal{C}_v$ as a dissimilarity score, to find the most dissimilar node $u^*$ for connecting. The candidate set $\mathcal{C}_v \subset \mathcal{V}$ of size $k$ (default $k=4096$) is randomly sampled from the remaining nodes. The dissimilarity score and the newly added edge are defined as:
\begin{equation}
    d(v,u) = \|\mathbf{x}_v - \mathbf{x}_u\|_2, \forall u \in \mathcal{C}_v ,
\end{equation}
\begin{equation}
\Delta \mathcal{E}_v = \{(u^*, v) | u^* = \arg\!\max_{u \in \mathcal{C}_v} d(v,u)\}.
\end{equation}

\paragraph{\textbf{Structural Reorganization Anomaly ($\tau_3$)}}
This anomaly preserves the same degree as its previous behavior while the local neighborhood structure is completely reorganized \cite{wang2019detecting}. In communication networks, this could represent the behaviors of compromised accounts that maintain their typical activity level but suddenly change all their communication partners.

Specifically, we remove all the existing edges from the target node $v$ and randomly generate the same number of new edges. This process can be formulated as follows:
\begin{equation}
\mathcal{E}' = \mathcal{E} \setminus \{(u,v) | u \in \mathcal{N}(v)\},
\end{equation}
\begin{equation}
\Delta \mathcal{E}_v = \{(u_i,v) | u_i \sim \text{Uniform}(\mathcal{V}), i \in [1,d_v]\},
\end{equation}
where $\mathcal{N}(v)$ represents the original neighbors of node $v$, and $d_v$ is the original degree of the target node.

\paragraph{\textbf{Feature Replacement Anomaly ($\tau_4$)}}
This anomaly maintains structural pattern while maximizing feature deviation \cite{song2007conditional}. This corresponds to situations where a user account in an online platform maintains its friendship network but exhibits significantly different content-sharing behavior, indicating account hijacking.

Specifically, we replace the target node's features with those of the most dissimilar node sampled from a candidate set $\mathcal{C}_v$, while preserving all local structural connections. The candidate set $\mathcal{C}_v$ is constructed the same way as in $\tau_2$. The replacement $\mathbf{x}'_v$ for the target node can be expressed as:
\begin{equation}
\mathbf{x}'_v = \mathbf{x}_{u^*}, \text{ where } u^* = \arg\!\max_{u \in \mathcal{C}_v} \|\mathbf{x}_v - \mathbf{x}_u\|_2.
\end{equation}

\paragraph{\textbf{Feature Perturbation ($\tau_5$)}}
This generates subtle feature anomalies by perturbing a subset of feature dimensions. In IoT sensor networks, this may correspond to the situation of malfunctioning sensors reporting higher readings in specific dimensions, or in healthcare networks, patients showing subtle deviations across a few vital signs that together indicate an emerging condition.

Specifically, we introduce controlled perturbations to a small subset of feature dimensions while keeping the rest unchanged. The modifications of node features for dimension $i$ are expressed as:
\begin{equation}
\mathbf{x}'_{v,i} =
\begin{cases}
\mathbf{x}_{v,i} + \beta \cdot \sigma_i, & \text{if } i \in \mathcal{S} \\
\mathbf{x}_{v,i}, & \text{otherwise}
\end{cases},
\end{equation}
where $\mathcal{S} \subset \{1,...,d\}$ is a small subset of feature dimensions randomly sampled,
$\beta \sim \text{Uniform}(3, 5)$ is a scaling factor and $\sigma_i = \sqrt{\frac{1}{n}\sum_{j=1}^n (\mathbf{x}_{j,i} - \bar{\mathbf{x}}_i)^2}$ is the standard deviation of the $i$-th feature across all nodes.

\subsubsection{Feature Encoding with Multi-task Learning}

The feature encoder conducts a multi-task learning as an auxiliary task to make the latent feature representation be sensitive to different types of disturbances in graph, leveraging the aforementioned generated synthetic anomalies. This is different from traditional self-supervised learning methods designed for general representation learning. The traditional methods aim to capture inherent patterns in data. In contrast, our method is specifically designed for anomaly detection. In particular, a GNN is employed to generate the latent representation, followed by a multi-task learning process for optimization. It is expected that the anomaly detector learned on the representation can be discriminant for diverse types of abnormal patterns.

The component includes a shared GNN encoder, followed by specialized detection heads for each anomaly type. We choose GATSep~\cite{platonov2023gatsep} as our encoder due to its effectiveness in capturing node interactions under heterophilious conditions.
As shown in \cref{fig:overview}, a synthetic anomaly-enhanced graph incorporates multiple types of perturbations. Each perturbation operation $\tau_k$ (where $k \in \{1,...,K\}$ and $K$ represents the total number of synthetic anomaly types) is applied to a distinct set of $s$ nodes randomly sampled from the unlabeled node set $\mathcal{U}$. The generation process can be formalized as:
\begin{equation}
\mathcal{V}_k \sim \text{Uniform}(\mathcal{U}, s),
\end{equation}
\begin{equation}
\mathcal{G}'_0 = \mathcal{G},
\end{equation}
\begin{equation}
\mathcal{G}'_k = \tau_k(\mathcal{G}'_{k-1}, \mathcal{V}_k), \quad k \in \{1,...,K\}.
\end{equation}

The synthetic anomaly-enhanced graph $\mathcal{G}' = \mathcal{G}'_K$ is then fed into the GNN encoder to produce node-level embedding:
\begin{equation}
\mathbf{H}' = \text{GNN}(\mathcal{G}') \in \mathbb{R}^{n \times d'},
\end{equation}
where $\mathbf{H}'$ represents the node embedding matrix and $d'$ is the embedding dimension.

The output of the GNN is then optimized through a multi-task learning process\cite{zhang2021survey}. We design five specialized detection heads, each corresponding to one type of synthetic anomalies. These heads are implemented as two-layer perceptrons that perform binary classification:
\begin{equation}
p_k(v) = \sigma(\text{MLP}_k(\mathbf{h}'_v)), \quad k \in \{1,...,K\},
\end{equation}
where $\mathbf{h}'_v \in \mathbf{H}'$ is the embedding for node $v$, $\sigma(\cdot)$ is the sigmoid activation function, and $p_k(v) \in [0,1]$ represents the probability of node $v$ being a synthetic anomaly of type $k$.

\subsection{Anomaly Detector}

An anomaly detector generates the probability of each input sample being anomalous, through a weakly supervised learning with limited number of labeled anomalies. While utilizing a dedicated detection head for real anomalies, the detector leverages the same GNN encoder that has been trained to recognize various synthetic perturbation patterns, enabling it to benefit from the rich feature representations learned through the multi-task learning process.

The detection process can be represented as:
\begin{equation}
\mathbf{H} = \text{GNN}(\mathcal{G}),
\end{equation}
\begin{equation}
f(v) = \sigma(\text{MLP}_{\text{score}}(\mathbf{h}_v)),
\end{equation}
where $\mathbf{h}_v \in \mathbf{H}$ is the embedding for node $v$, $\sigma(\cdot)$ is the sigmoid activation function, and $f(v) \in [0,1]$ represents the estimated probability of node $v$ being an anomaly.

\subsection{Objectives and Training Procedure}

Our model aims to detect anomalous nodes based on a meaningful feature representation that is guided by synthetically generated anomalies. The core of the entire training includes two aspects: (a) learn a disturbance-sensitive feature representation that is attuned to diverse types of abnormal patterns in graphs; (b) enhance the latent representation with domain knowledge from real graph anomalies. 

The aforementioned goal is achieved through a carefully designed loss function that includes two complementary parts. The joint loss can be represented as: 
\begin{equation}
\mathcal{L} = \mathcal{L}_{\text{real}} + \lambda\mathcal{L}_{\text{synth}} \label{eq:main_loss},
\end{equation}
where $\lambda$ is a hypermeter that controls the balance between the two parts of loss, \textit{i.e.}, the auxiliary task loss $\mathcal{L}_{\text{synth}}$ and the real anomaly supervised loss $\mathcal{L}_{\text{real}}$.

The $\mathcal{L}_{\text{synth}}$ is to help the model learn a latent feature representation that can separate normal samples from arbitrary types of synthetic perturbations. We expect that the learned latent representation is sensitive to diverse types of disturbances. This gives us the following accumulation term:
\begin{equation}
\mathcal{L}_{\text{synth}} = \sum_{k=1}^K \mathcal{L}_k. \label{eq:l_synth}
\end{equation}

Each individual loss $\mathcal{L}_k$ is defined as:
\begin{equation}
\mathcal{L}_k = -\frac{1}{|\mathcal{B}_k|}\sum_{v \in \mathcal{B}_k} [y_v^k \log(p_k(v)) + (1-y_v^k)\log(1-p_k(v))],
\end{equation}
where $\mathcal{B}_k$ represents the current batch containing equal numbers of specific perturbed and normal nodes, and $y_v^k$ indicates whether node $v$ was perturbed using the $k$-th perturbation type.

The $\mathcal{L}_{\text{real}}$ further enhances the latent representation of model to distinguish real anomalies from the normal ones, which gives the following loss term:
\begin{equation}
\mathcal{L}_{\text{real}} = -\frac{1}{|\mathcal{B}|}\sum_{v \in \mathcal{B}} [y_v \log(f(v)) + (1-y_v)\log(1-f(v))], \label{eq:l_real}
\end{equation}
where each batch $\mathcal{B}$ contains an equal number of normal and anomalous nodes sampled from the training set.

A two-phase procedure is designed to train the proposed model. Instead of directly training with the combined objective $\mathcal{L}$ from scratch, our training procedure introduces a warm-up phase that only employs synthetic perturbations to develop disturbance-sensitive
latent representation, without any labeled anomalies. This warm-up phase is necessary because directly training with limited labeled anomalies is prone to overfitting the model. From empirical study, we also observe the beneficial effects of warm-up, facilitating the detection model to be away from suboptimal performance. The optimization objective of this phase is then treated as a regularization term in the loss function of the next full-training phase, which incorporates supervision signals to enhance the detection performance. The aforementioned considerations motivate our two-phase training procedure:
\begin{itemize}
    \item A warm-up phase to develop latent representation that is sensitive to various graph perturbations using only $\mathcal{L}_{\text{synth}}$;
    \item A full-training phase that incorporates perturbation learning with supervision  of real anomalies using $\mathcal{L}$.
\end{itemize}

Details of the training procedure are demonstrated in Algorithm~\ref{algo:train}. Note that in the two-phase training, the GNN module parameters in the full-training phase are inherited from the warm-up phase rather than reinitialized. 

\begin{algorithm}[t]
\caption{Two-Phase Training with Synthetic Anomalies}
\label{algo:train}
\begin{algorithmic}[1]
\State Initialize model parameters $\Theta$

\Statex \textbf{Warm-up Phase}
\For{epoch = 1 to $N_{warm}$}
    \State $\mathcal{G}'_0 \leftarrow \mathcal{G}$
    \For{k = 1 to K}
        \State $\mathcal{V}_k \leftarrow \text{Sample}(\mathcal{U}, s)$
        \State $\mathcal{G}'_k \leftarrow \tau_k(\mathcal{G}'_{k-1}, \mathcal{V}_k)$
        \State {\small $\mathcal{B}_k \leftarrow \text{ConstructBatch}(\mathcal{G}'_k)$, $y_k \leftarrow \text{CreateLabels}(\mathcal{V}_k)$}
    \EndFor
        \State $\mathbf{H}' \leftarrow \text{GNN}(\mathcal{G}'_K)$
        \State $\mathcal{L}_{\text{synth}} \leftarrow \sum_{k=1}^K \text{BCELoss}(\text{MLP}_k(\mathbf{H}'[\mathcal{B}_k]), y_k)$
        \State Update $\Theta$ using $\nabla\mathcal{L}_{\text{synth}}$
\EndFor

\Statex \textbf{Full-training Phase}
\For{epoch = 1 to $N$}
    \State $\mathcal{B} \sim \text{BalancedBatch}(\mathcal{G})$
    \State $\mathcal{L}_{\text{real}} \leftarrow \text{BCELoss}(\text{MLP}(\text{GNN}(\mathcal{G})[\mathcal{B}]), y[\mathcal{B}])$
    \State Calculate $\mathcal{L}_{\text{synth}}$ as in Warm-up Phase
    \State Update $\Theta$ using $\nabla(\mathcal{L}_{\text{real}} + \lambda\mathcal{L}_{\text{synth}})$
\EndFor
\end{algorithmic}
\end{algorithm}

\section{Experiments}\label{sec:exp}

We conduct extensive experiments to evaluate our proposed method on real-world graph datasets. Through these experiments, we aim to validate the effectiveness of our approach and analyze the contribution of each component to the model's performance.

\subsection{Experimental Setup}

\subsubsection{Datasets}

We evaluate our method on five real-world graph datasets, \textit{i.e.}, Amazon\cite{mcauley2013amateurs}, Yelp\cite{rayana2015collective}, Weibo\cite{kumar2019predicting}, Questions\cite{platonov2023gatsep} and T-Finance\cite{tang2022bwgnn}, which span different application domains. All the labeled anomalous nodes in the five datasets are realistic anomalies, such as product review fraud (Amazon), business reviews fraud (Yelp), spam on social networks (Weibo), suspicious activities (Questions) and financial fraud (T-Finance). The statistics of the five datasets are summarized in Table~\ref{tab:datasets}. Detailed descriptions of datasets are as follows:
\begin{itemize}
    \item The \textbf{Amazon}\cite{mcauley2013amateurs} dataset focuses on identifying users paid to write fake reviews for products in the Musical Instrument category on Amazon.com. Users are connected through shared product reviews, ratings, or high review similarities.
    
    \item The \textbf{Yelp}\cite{rayana2015collective} dataset is designed to detect anomalous reviews that unfairly promote or demote products on Yelp.com. The dataset includes complex edge types such as reviews by the same user, reviews for the same product with identical ratings, and reviews posted in the same month.
    
    \item The \textbf{Weibo}\cite{kumar2019predicting} dataset represents a social network from the Weibo platform. The anomalies are identified based on suspicious activities, specifically posts made within short timeframes. Users are labeled as suspicious based on their engagements at least in five specific temporal activities.
    
    \item The \textbf{Questions}\cite{platonov2023gatsep} dataset, collected from Yandex Q, focuses on users interested in the medical domain. The graph connects users who interact through question-answering over a one-year period. Nodes are labeled as anomalous based on their sustained platform activity.

    \item The \textbf{T-Finance}\cite{tang2022bwgnn} dataset aims to identify anomalous accounts in transaction networks. Nodes represent anonymized accounts with features related to registration information, logging activities, and interaction frequency. Edges connect accounts with transaction records. Nodes are labeled as anomalous if they fall into categories like fraud, money laundering, and online gambling.
\end{itemize}

\begin{table}[t]
\centering
\caption{Statistics for the Five Datasets.}
\label{tab:datasets}
\renewcommand{\arraystretch}{1.2}
\setlength{\tabcolsep}{2pt}
\begin{tabular}{lccccc}
\toprule
\textbf{Dataset} & \textbf{Nodes} & \textbf{Edges} & \textbf{Anomalies} & \textbf{\%Anom.} & \textbf{Domain} \\
\midrule
Amazon     & 11,944  & 4,398,392  & 821   & 6.87\%  & Reviews \\
Yelp       & 45,954  & 3,846,979  & 6,677 & 14.53\% & Reviews \\
Weibo      & 8,405   & 407,963    & 868   & 10.33\% & Social Media \\
Questions  & 48,921  & 153,540    & 1,460 & 2.98\%  & Q\&A Platform \\
T-Finance  & 39,357  & 21,222,543 & 1,803 & 4.58\%  & Finance \\
\bottomrule
\end{tabular}
\end{table}

\subsubsection{Competing Methods}

To evaluate the performance of the proposed method, we compare it with two categories of methods, \textit{i.e.}, graph anomaly detection methods that utilize the labeled anomalies available and baseline methods that directly employ the binary classification loss to train popular deep neural networks for graph.
Specifically, seven methods, \textit{i.e.}, PC-GNN\cite{liu2021pcgnn}, 
H2-FDetector\cite{shi2022h2},
BWGNN\cite{tang2022bwgnn},
GHRN\cite{gao2023ghrn}, ConsisGAD\cite{chen2024consisgad}, 
CARE-GNN\cite{dou2020caregnn},
and GATSep\cite{platonov2023gatsep}, are included in the first category. Note that we modified GATSep with an alternative output head, as in ~\eqref{eq:l_real}, for graph anomaly detection. For the baseline methods, MLP, GCN\cite{kipf2017gcn}, GraphSAGE\cite{hamilton2017graphsage}, GAT\cite{veličković2018gat}, 
GIN\cite{xu2018gin} are utilized as the network structure, respectively.  

\subsubsection{Experimental Settings}
For baseline methods that directly adopt MLP or GNNs as the network structure, we employ a consistent architecture across all experiments. The MLP based baseline consists of two layers with a dimension of 64 for hidden units. For GNN based baselines (GCN, GraphSAGE, GAT, and GIN), we use a two-layer GNN with 64-dimensional node embeddings, followed by a two-layer MLP classifier with hidden dimension of 64. The GNN encoder of the proposed method consists of a two-layer GATSep with output dimensions of 64, followed by a two-layer MLP with hidden dimension of 64. ReLU activation functions are applied after each layer in the above methods. For specialized graph anomaly detection methods, we utilize their published implementations with their recommended configurations. 

For the settings of the proposed method, the number of synthetic samples per type ($s$) is set to 32, which determines the batch size for synthetic anomalies. For real anomalies, we use a batch size of 512. In the context of synthetic anomaly generation, we sampled $|\mathcal{S}| \sim \text{Uniform}(2, \min(5, 0.1d))$ to ensure that only a small subset of feature dimensions is perturbed. The model is optimized using Adam optimizer~\cite{kingma2014adam} with default PyTorch parameters ($\beta_1=0.9$, $\beta_2=0.999$, $\epsilon=1e-8$). To mitigate overfitting in the limited labeled data setting, we incorporate an $l_2$-norm regularizer with weight set to 0.01. The trade-off parameter $\lambda$ in the joint loss function is set to 20 for the Yelp dataset and 4 for all other datasets.

Our implementation uses Python 3.12.4, PyTorch 2.3.1, Deep Graph Library (DGL) 2.3.0, and scikit-learn 1.4.2. All experiments were conducted on a machine equipped with an Intel Core i9-13900K processor, NVIDIA RTX 4090 GPU with 24GB VRAM, and 128GB system memory.

\subsubsection{Evaluation Metrics}

In the evaluation process, we employ two widely recognized performance indicators: the Area Under the Receiver Operating Characteristic Curve (AUROC) and the Area Under the Precision-Recall Curve (AUPRC). Higher values in both metrics indicate superior performance.

\subsection{Performance Comparison with Competing Methods}
\label{sec:main_compare}

\begin{table*}[t]
\centering
\caption{Experimental Results (Mean $\pm$ Standard Deviation) of Our Proposed Method and Competing Methods. The Best Result on Each Dataset Is in \textbf{Bold}, and the Second-Best Result Is \underline{Underlined}. OOM denotes out of memory on a 24GB GPU.OOM denotes out of memory on a 24GB GPU.}
\label{tab:comparison_results}
\renewcommand{\arraystretch}{1.3}
\setlength{\tabcolsep}{2pt}
\begin{tabular}{lcccccccccccc}
\toprule
\multirow{2}{*}{\textbf{Model}} & \multicolumn{6}{c}{\textbf{AUROC}} & \multicolumn{6}{c}{\textbf{AUPRC}} \\
\cmidrule(lr){2-7} \cmidrule(lr){8-13}
& Amazon & Yelp & Weibo & Questions & T-Finance & \begin{tabular}{c}Avg. \\ Rank\end{tabular} & Amazon & Yelp & Weibo & Questions & T-Finance & \begin{tabular}{c}Avg. \\ Rank\end{tabular} \\
\midrule
MLP & 0.903{\scriptsize $\pm$0.01} & 0.712{\scriptsize $\pm$0.02} & 0.718{\scriptsize $\pm$0.03} & 0.609{\scriptsize $\pm$0.03} & 0.882{\scriptsize $\pm$0.02} & 9.20 & 0.818{\scriptsize $\pm$0.02} & 0.679{\scriptsize $\pm$0.03} & 0.738{\scriptsize $\pm$0.04} & 0.207{\scriptsize $\pm$0.02} & 0.758{\scriptsize $\pm$0.04} & 9.00 \\
GCN\cite{kipf2017gcn} & 0.822{\scriptsize $\pm$0.00} & 0.523{\scriptsize $\pm$0.04} & 0.930{\scriptsize $\pm$0.07} & 0.473{\scriptsize $\pm$0.08} & 0.917{\scriptsize $\pm$0.00} & 10.60 & 0.582{\scriptsize $\pm$0.03} & 0.475{\scriptsize $\pm$0.04} & 0.898{\scriptsize $\pm$0.13} & 0.132{\scriptsize $\pm$0.02} & 0.840{\scriptsize $\pm$0.00} & 11.00 \\
GraphSAGE\cite{hamilton2017graphsage} & 0.894{\scriptsize $\pm$0.01} & 0.693{\scriptsize $\pm$0.03} & 0.935{\scriptsize $\pm$0.04} & 0.502{\scriptsize $\pm$0.09} & 0.892{\scriptsize $\pm$0.01} & 9.60 & 0.798{\scriptsize $\pm$0.02} & 0.653{\scriptsize $\pm$0.03} & 0.912{\scriptsize $\pm$0.08} & 0.150{\scriptsize $\pm$0.04} & 0.802{\scriptsize $\pm$0.01} & 9.60 \\
GAT\cite{veličković2018gat} & 0.918{\scriptsize $\pm$0.01} & 0.722{\scriptsize $\pm$0.01} & 0.869{\scriptsize $\pm$0.11} & 0.604{\scriptsize $\pm$0.02} & 0.917{\scriptsize $\pm$0.01} & 5.80 & 0.857{\scriptsize $\pm$0.02} & 0.691{\scriptsize $\pm$0.02} & 0.856{\scriptsize $\pm$0.11} & 0.185{\scriptsize $\pm$0.02} & 0.876{\scriptsize $\pm$0.01} & 5.20 \\
GIN\cite{xu2018gin} & 0.910{\scriptsize $\pm$0.01} & 0.700{\scriptsize $\pm$0.05} & 0.913{\scriptsize $\pm$0.06} & 0.589{\scriptsize $\pm$0.04} & 0.914{\scriptsize $\pm$0.00} & 7.60 & 0.856{\scriptsize $\pm$0.01} & 0.649{\scriptsize $\pm$0.06} & 0.894{\scriptsize $\pm$0.05} & 0.172{\scriptsize $\pm$0.01} & 0.867{\scriptsize $\pm$0.00} & 7.60 \\
\midrule
PCGNN\cite{liu2021pcgnn} & 0.907{\scriptsize $\pm$0.01} & 0.727{\scriptsize $\pm$0.02} & 0.690{\scriptsize $\pm$0.09} & 0.576{\scriptsize $\pm$0.02} & \underline{0.926{\scriptsize $\pm$0.00}} & 7.20 & 0.785{\scriptsize $\pm$0.03} & 0.695{\scriptsize $\pm$0.02} & 0.652{\scriptsize $\pm$0.14} & 0.200{\scriptsize $\pm$0.01} & 0.868{\scriptsize $\pm$0.01} & 7.80 \\
H2FD\cite{shi2022h2} & 0.870{\scriptsize $\pm$0.01} & 0.674{\scriptsize $\pm$0.02} & 0.813{\scriptsize $\pm$0.11} & 0.565{\scriptsize $\pm$0.01} & OOM & 11.50 & 0.736{\scriptsize $\pm$0.02} & 0.650{\scriptsize $\pm$0.02} & 0.789{\scriptsize $\pm$0.12} & 0.152{\scriptsize $\pm$0.01} & OOM & 11.25 \\
BWGNN\cite{tang2022bwgnn} & 0.885{\scriptsize $\pm$0.02} & 0.718{\scriptsize $\pm$0.02} & 0.821{\scriptsize $\pm$0.04} & 0.611{\scriptsize $\pm$0.03} & 0.921{\scriptsize $\pm$0.01} & 6.80 & 0.845{\scriptsize $\pm$0.01} & 0.685{\scriptsize $\pm$0.03} & 0.810{\scriptsize $\pm$0.04} & 0.242{\scriptsize $\pm$0.02} & 0.860{\scriptsize $\pm$0.01} & 6.40 \\
GHRN\cite{gao2023ghrn} & 0.885{\scriptsize $\pm$0.02} & 0.725{\scriptsize $\pm$0.02} & 0.795{\scriptsize $\pm$0.05} & 0.605{\scriptsize $\pm$0.02} & 0.923{\scriptsize $\pm$0.01} & 7.20 & 0.842{\scriptsize $\pm$0.01} & 0.691{\scriptsize $\pm$0.02} & 0.778{\scriptsize $\pm$0.07} & 0.244{\scriptsize $\pm$0.01} & 0.861{\scriptsize $\pm$0.01} & 6.40 \\
ConsisGAD\cite{chen2024consisgad} & \underline{0.952{\scriptsize $\pm$0.01}} & 0.716{\scriptsize $\pm$0.02} & 0.730{\scriptsize $\pm$0.03} & \underline{0.638{\scriptsize $\pm$0.06}} & OOM & 5.75 & \underline{0.917{\scriptsize $\pm$0.01}} & 0.674{\scriptsize $\pm$0.03} & 0.643{\scriptsize $\pm$0.07} & \textbf{0.268{\scriptsize $\pm$0.03}} & OOM & 6.25 \\
CAREGNN\cite{dou2020caregnn} & 0.899{\scriptsize $\pm$0.01} & \textbf{0.848{\scriptsize $\pm$0.04}} & 0.861{\scriptsize $\pm$0.00} & 0.606{\scriptsize $\pm$0.01} & 0.910{\scriptsize $\pm$0.00} & 6.60 & 0.813{\scriptsize $\pm$0.01} & \textbf{0.826{\scriptsize $\pm$0.03}} & 0.853{\scriptsize $\pm$0.00} & 0.185{\scriptsize $\pm$0.01} & 0.845{\scriptsize $\pm$0.00} & 7.20 \\
GATSep\cite{platonov2023gatsep} & 0.937{\scriptsize $\pm$0.01} & 0.722{\scriptsize $\pm$0.02} & \underline{0.947{\scriptsize $\pm$0.01}} & 0.631{\scriptsize $\pm$0.01} & 0.919{\scriptsize $\pm$0.03} & \underline{3.80} & 0.895{\scriptsize $\pm$0.02} & 0.676{\scriptsize $\pm$0.03} & \underline{0.926{\scriptsize $\pm$0.01}} & 0.218{\scriptsize $\pm$0.01} & \underline{0.879{\scriptsize $\pm$0.03}} & \underline{4.20} \\
CGNN\cite{li2025cgnn} & 0.909{\scriptsize $\pm$0.02} & 0.523{\scriptsize $\pm$0.02} & 0.665{\scriptsize $\pm$0.02} & 0.595{\scriptsize $\pm$0.06} & 0.911{\scriptsize $\pm$0.02} & 10.20 & 0.804{\scriptsize $\pm$0.03} & 0.451{\scriptsize $\pm$0.03} & 0.526{\scriptsize $\pm$0.04} & \underline{0.265{\scriptsize $\pm$0.04}} & 0.864{\scriptsize $\pm$0.03} & 9.20 \\
\midrule
\textbf{Ours} & \textbf{0.969{\scriptsize $\pm$0.01}} & \underline{0.755{\scriptsize $\pm$0.02}} & \textbf{0.962{\scriptsize $\pm$0.01}} & \textbf{0.657{\scriptsize $\pm$0.02}} & \textbf{0.932{\scriptsize $\pm$0.02}} & \textbf{1.20} & \textbf{0.941{\scriptsize $\pm$0.01}} & \underline{0.719{\scriptsize $\pm$0.02}} & \textbf{0.936{\scriptsize $\pm$0.01}} & 0.237{\scriptsize $\pm$0.01} & \textbf{0.887{\scriptsize $\pm$0.03}} & \textbf{2.00} \\
\bottomrule
\end{tabular}
\end{table*}

\paragraph{Experiment Settings}

We evaluate the proposed method with the competing methods on four real-world datasets. For fair comparison, we establish consistent hyper-parameter settings across different methods, \textit{i.e.}, all the hidden dimensions in MLPs and GNNs are 64, if not specified. For specialized graph anomaly detection methods, we use the published implementations and configurations. In all experiments, we split each dataset with a proportion of 8:1:1, w.r.t., the training, validation and test sets. 
Our experimental setup follows established practices from weakly supervised anomaly detection literature (\cite{pang2023prenet, zhou2021feawad, pang2019devnet}), where there are a fixed amount of anomalous labeled samples and unlabeled samples with contamination available during training.
We retain only $m = 30$ labeled anomalies in the training set, which only account for 0.16\%-0.9\% of the training data. To simulate real-world scenarios where unlabeled data usually contain anomalous samples,
we add anomalies, randomly sampled from the remaining labeled anomalies in the training set, to normal class at a contamination rate of 1\%. The total number of unlabeled samples in the training set remains fixed across different $m$ values. The only change is the number of anomaly labels available—as $m$ increases, more anomalies are explicitly labeled.
We report average performance across 16 runs to ensure statistical robustness and minimize the impact of random initialization. $\lambda$ in our method is set to 20 for Yelp while it is denoted as 4 for the other three datasets. 

\paragraph{Result Analysis}
\cref{tab:comparison_results} presents the comparative performance between our method and the competing methods. The results demonstrate that our method consistently outperforms baseline methods directly training with MLP or GNNs, across all datasets on both AUROC and AUPRC metrics. Among methods specifically designed for graph anomaly detection (GAD), our method achieves superior overall performance. In terms of AUROC, our method attains the highest performance on three of all the four datasets. It ranks second on Yelp dataset behind CAREGNN, which is specifically designed to handle such camouflaged anomalies through specialized neighbor selection. However, our method substantially outperforms CAREGNN on all other datasets. Similar performance improvements are observed for AUPRC. For the Questions dataset, our AUPRC is lower than ConsisGAD due to its specific optimization objective (AUPRC) in the training process. Overall, our method achieves an average rank of 1.20 for AUROC and 2.00 for AUPRC, outperforming all competing methods. These significant performance gains indicate the benefits of our feature encoding module and the specific designed loss function, which are the major distinction between our method and existing GAD methods, highlighting the crucial role of our novel encoding strategy. 

Notably, the performance of GAD methods sometimes falls below baseline methods when putting into the weakly-supervised setting with datasets involving real anomalies. We hypothesize that the labeled anomalies under realistic scenarios as well as larger model parameters tend to over-fit the GAD methods.

\subsection{Sample Efficiency}
\label{sec:sample_efficiency}

\paragraph{Experiment Settings}
To investigate the sample efficiency of the proposed method, we evaluate the performance of 
different graph anomaly detection methods with various number of available labeled samples. For all the evaluated methods, the number of available labeled samples is set to 8, 15, 30, 60 and 120 respectively. The rest of experiment settings are the same as in Section~\ref{sec:main_compare}.

\begin{figure*}[t]
\centering
\includegraphics[width=1\linewidth]{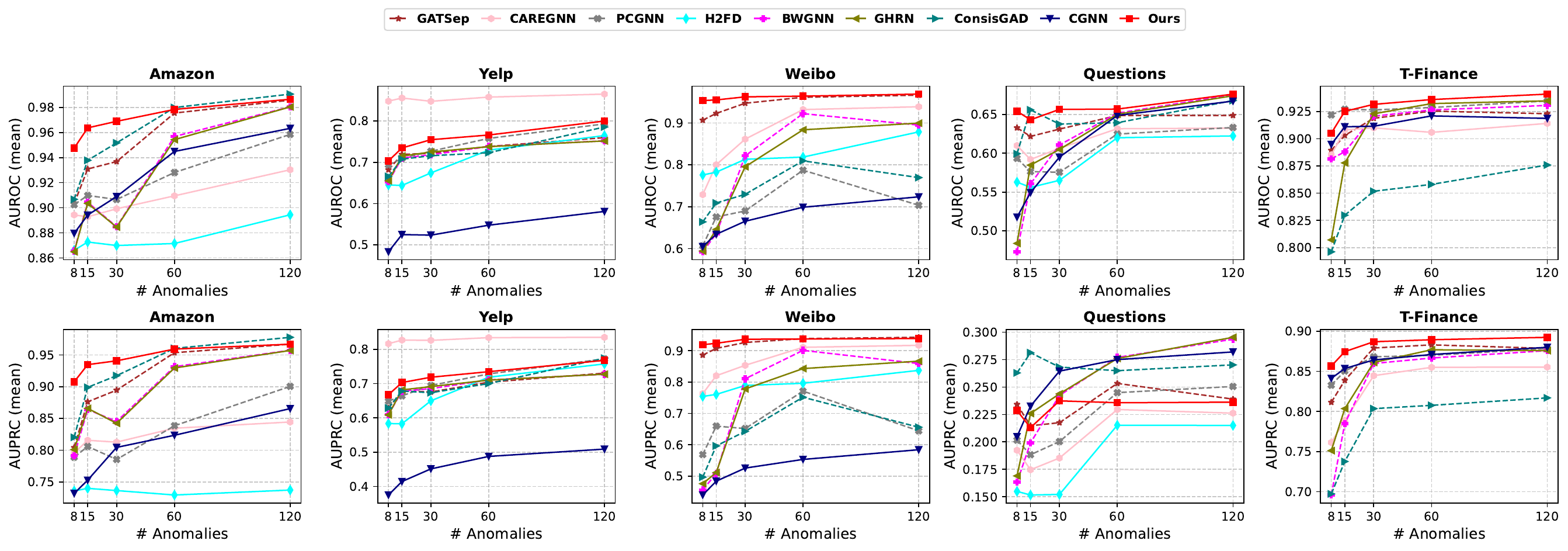}
\caption{AUROC and AUPRC performance w.r.t. the number of labeled anomalies}
\label{fig:anomaly_num_full}
\end{figure*}

\paragraph{Result Analysis}
\cref{fig:anomaly_num_full} presents the performance comparisons between our method and other GAD methods under varying number of labeled samples. The experimental results demonstrate that all methods generally show improved performance with increased label availability. However, the detection performance of the competing methods sometimes degrades when more labeled data are available. This might be due to the variety of anomaly types that usually increases with the number of anomalies, resulting the decreased performance from those conflicting information. In contrast, our method shows stable improvements as the number of labeled anomalies increases. 

Our method performs particularly well when few labeled anomalies are available, \textit{i.e.}, 8, 15, and 30. It achieves the best average performance on 3 of 4 datasets among the other GAD methods, except for Yelp dataset, where our method ranks the second as explained in Section~\ref{sec:main_compare}.
Moreover, our method is the most sample efficient method, requiring much less labeled anomalies to achieve competitive performance than other methods.
Specifically, our method needs only a fraction of labeled samples to achieve the same results as the best competing method, \textit{e.g.}, our method requires 69\% less samples than the best competing method ConsisGAD on Amazon, and achieves comparable performance as the the second ranking method GATSep on Weibo using only 82\% less labeled data.
This superior performance is due to the effective feature representation in our method which leverages synthetic generated anomalies to guide disturbance-sensitive feature learning,
resulting in better efficiency than the counterpart graph anomaly detection methods.

\subsection{Ablation Study}
\label{sec:ablation}

In this section, we perform ablation studies to explore the effectiveness of different components of our framework. Note that, unless stated otherwise, the experimental settings are the same as in Section~\ref{sec:main_compare}.

\begin{table*}[t]
\centering
\caption{Experimental Results for Evaluating the Beneficial Effect of Each Component in the Training Procedure.}
\label{tab:ablation_phase1_and_regularization}
\renewcommand{\arraystretch}{1.2}
\setlength{\tabcolsep}{1.5pt}
\begin{tabular}{cccccccccccc}
\toprule
\multirow{2}{*}{\textbf{Warm-up}} & \multirow{2}{*}{\textbf{Regularize}} & \multicolumn{5}{c}{\textbf{AUROC}} & \multicolumn{5}{c}{\textbf{AUPRC}} \\
\cmidrule(lr){3-7} \cmidrule(lr){8-12}
& & Amazon & Yelp & Weibo & Questions & T-Finance & Amazon & Yelp & Weibo & Questions & T-Finance \\
\midrule
 &  & 0.937±0.012 & 0.722±0.017 & 0.947±0.011 & 0.631±0.005 & 0.919±0.034 & 0.895±0.020 & 0.676±0.026 & 0.926±0.014 & 0.218±0.013 & 0.879±0.030 \\
 & \checkmark & 0.968±0.006 & 0.726±0.014 & 0.959±0.007 & 0.641±0.010 & 0.925±0.019 & 0.941±0.010 & 0.692±0.020 & 0.930±0.013 & 0.226±0.013 & 0.881±0.026 \\
\checkmark &  & 0.944±0.010 & 0.727±0.019 & 0.953±0.021 & 0.637±0.008 & 0.921±0.021 & 0.900±0.019 & 0.691±0.022 & 0.921±0.053 & 0.229±0.009 & 0.878±0.031 \\
\checkmark & \checkmark & \textbf{0.969}±0.005 & \textbf{0.755}±0.017 & \textbf{0.962}±0.008 & \textbf{0.657}±0.017 & \textbf{0.932}±0.022 & \textbf{0.941}±0.008 & \textbf{0.719}±0.023 & \textbf{0.936}±0.011 & \textbf{0.237}±0.011 & \textbf{0.887}±0.034 \\
\bottomrule
\end{tabular}
\end{table*}

\paragraph{Effect of Objective Design and Training Procedure}
To evaluate the effectiveness of our two-phase training procedure, we construct two variants of the proposed method. The first variant (\textit{w/o warm-up}) removes the warm-up phase and directly trains the model with $\mathcal{L}$ from scratch instead. The second variant (\textit{w/o regularization}) retains the warm-up phase but only use $\mathcal{L}_{\text{real}}$ during the second training phase. We also include GATSep as a baseline, which can be viewed as our method without both warm-up and regularization. Through these variants, we aim to understand: (1) the importance of the warm-up phase in establishing a discriminative feature space guided by synthetic anomalies; (2) the role of synthetic anomaly regularization in preserving learned abnormal patterns during the full training phase.

The experiment results in \cref{tab:ablation_phase1_and_regularization} indicates the importance of both warm-up and regularization in our framework. The \textit{w/o warm-up} variant shows an average performance decrease of 1.7\% in AUROC and 1.6\% in AUPRC, highlighting the essential role of warm-up in developing a disturbance-sensitive feature space before training on real anomalies. The performance degradation becomes even more severe with the \textit{w/o regularization} variant, showing a 2.6\% decrease in AUROC and a 3.2\% decrease in AUPRC. This substantial drop indicates that without synthetic perturbation regularization in the full training phase, the model could not maintain the ability of being sensitive to various graph perturbations. Notably, both variants still outperform the baseline GATSep by 1.7\% and 0.8\% in AUROC respectively. Similar improvements can be observed in AUPRC. The above results demonstrate that even partial implementation of our training strategy contributes to improved anomaly detection.

\paragraph{Effect of Synthetic Anomalies}
To evaluate the effectiveness of synthetic anomalies, we create variants of our method by removing one specific type of synthetic anomaly at a time, respectively. We also include GATSep as a baseline, which is equivalent to our framework without any synthetic anomalies.

\begin{figure*}[ht]
\centering
\includegraphics[width=0.95\linewidth]{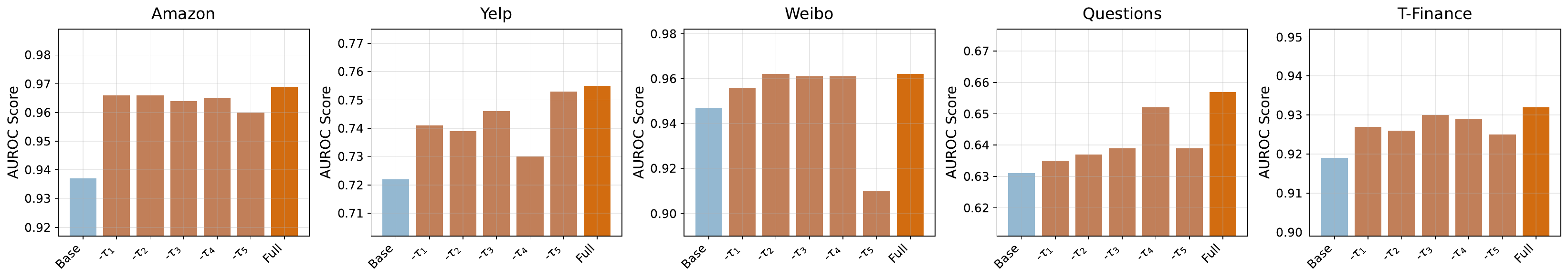}
\caption{AUROC results of experiments for evaluating the effect of synthetic anomalies.}
\label{fig:ablation_anomaly_types_auroc}
\end{figure*}

\begin{figure*}[ht]
\centering
\includegraphics[width=0.95\linewidth]{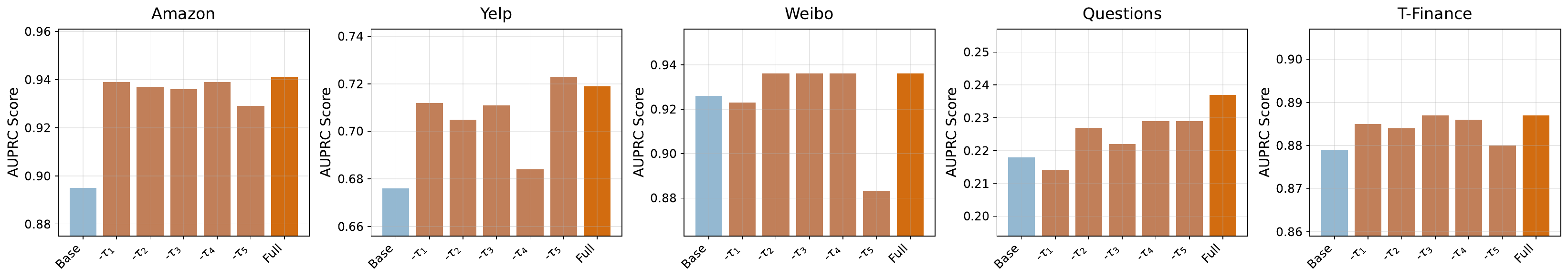}
\caption{AUPRC results of experiments for evaluating the effect of synthetic anomalies.} \label{fig:ablation_anomaly_types_auprc}
\end{figure*}

The experiment results in \cref{fig:ablation_anomaly_types_auroc} demonstrate the effectiveness of synthetic anomalies and their complementary contributions to the detection performance.
Our experiments reveal that the performance of the model degrades with any of the synthetic anomalies removed, falling behind the full model by 1.6\% on average in terms of AUROC. However, they still achieve better results than the baseline model, which does not use any of the synthetic anomalies, surpassing it by 1.9\% in AUROC on average. The full model, which incorporates all five types of synthetic perturbations, performs the best in all cases. As shown in \cref{fig:ablation_anomaly_types_auprc}, the AUPRC performance shows similar trends. These results demonstrate the value of our comprehensive synthetic perturbation choices. This validates the effectiveness of each synthetic anomaly design. The varying contribution levels reflect natural differences in dominant anomaly patterns across datasets, demonstrating appropriate adaptation through the learned discriminative ability to recognize perturbations.

\paragraph{Effectiveness of Feature Learning Module}
To evaluate the effectiveness of our disturbance-sensitive feature learning module, we compare it with the feature learning modules used in other graph anomaly detection methods. Specifically, we replace our disturbance-sensitive feature learning module with: (1) the autoencoder from DOMINANT\cite{ding2019dominant} that learns to reconstruct node features and graph structure; and (2) the contrastive learning module from CONAD\cite{xu2022conad} that learns node embeddings through contrastive design.

\begin{table*}[t]
\centering
\caption{Experimental Results for Evaluating the Effects of Different Feature Learning Techniques.}
\label{tab:ablation_self_supervised}
\renewcommand{\arraystretch}{1.2}
\setlength{\tabcolsep}{1.6pt}
\begin{tabular}{lcccccccccc}
\toprule
\multirow{2}{*}{\textbf{Configuration}} & \multicolumn{5}{c}{\textbf{AUROC}} & \multicolumn{5}{c}{\textbf{AUPRC}} \\
\cmidrule(lr){2-6} \cmidrule(lr){7-11}
& Amazon & Yelp & Weibo & Questions & T-Finance & Amazon & Yelp & Weibo & Questions & T-Finance \\
\midrule
Baseline (GATSep) & 0.937±0.012 & 0.722±0.017 & 0.947±0.011 & 0.631±0.005 & 0.919±0.034 & 0.895±0.020 & 0.676±0.026 & 0.926±0.014 & 0.218±0.013 & 0.879±0.030 \\
AE (DOMINANT) & 0.939±0.008 & 0.730±0.014 & 0.946±0.015 & 0.637±0.005 & 0.923±0.025 & 0.898±0.014 & 0.686±0.019 & 0.922±0.017 & 0.226±0.013 & 0.883±0.018 \\
Contrastive (CONAD) & 0.949±0.010 & 0.717±0.021 & 0.883±0.030 & 0.624±0.011 & 0.899±0.026 & 0.912±0.015 & 0.676±0.022 & 0.863±0.026 & 0.210±0.017 & 0.865±0.029 \\
\textbf{Ours} & \textbf{0.969}±0.005 & \textbf{0.755}±0.017 & \textbf{0.962}±0.008 & \textbf{0.657}±0.017 & \textbf{0.932}±0.022 & \textbf{0.941}±0.008 & \textbf{0.719}±0.023 & \textbf{0.936}±0.011 & \textbf{0.237}±0.011 & \textbf{0.887}±0.034 \\
\bottomrule
\end{tabular}
\end{table*}

As shown in \cref{tab:ablation_self_supervised}, the proposed method obtains substantial performance improvement in all cases. While both the autoencoder and contrastive learning approaches show varied performance compared to the baseline, our method achieves the most gains, outperforming the autoencoder-based approach by 2.8\% and the contrastive learning approach by 5.4\% in AUROC. In AUPRC, our method outperforms the autoencoder-based approach by 3.7\% and the contrastive learning approach by 6.5\%. These results validate that our domain-specific feature learning approach is more effective than existing feature learning techniques in the context of weakly supervised anomaly detection.

\paragraph{Effect of Specialized Detection Heads for Synthetic Anomalies}
To investigate the necessity of specialized detection heads for synthetic anomalies, we create a variant of our method without specialized heads. Specifically, The specialized detection heads are replaced with a single MLP head that employs the binary classification loss, treating all synthetic anomalies as a single anomalous class.  

\begin{table*}[t]
\centering
\caption{Experimental Results Without/With Specialized Detection Heads for Synthetic Anomalies.}
\label{tab:ablation_heads}
\renewcommand{\arraystretch}{1.2}
\setlength{\tabcolsep}{1.6pt}
\begin{tabular}{lcccccccccc}
\toprule
\multirow{2}{*}{\textbf{Configuration}} & \multicolumn{5}{c}{\textbf{AUROC}} & \multicolumn{5}{c}{\textbf{AUPRC}} \\
\cmidrule(lr){2-6} \cmidrule(lr){7-11}
& Amazon & Yelp & Weibo & Questions & T-Finance & Amazon & Yelp & Weibo & Questions & T-Finance \\
\midrule
w/o specialized heads & 0.962±0.007 & 0.737±0.023 & 0.958±0.006 & 0.654±0.020 & 0.928±0.039 & 0.932±0.012 & 0.710±0.022 & 0.933±0.012 & 0.236±0.018 & 0.884±0.032 \\
\textbf{Full} & \textbf{0.969}±0.005 & \textbf{0.755}±0.017 & \textbf{0.962}±0.008 & \textbf{0.657}±0.017 & \textbf{0.932}±0.022 & \textbf{0.941}±0.008 & \textbf{0.719}±0.023 & \textbf{0.936}±0.011 & \textbf{0.237}±0.011 & \textbf{0.887}±0.034 \\
\bottomrule
\end{tabular}
\end{table*}

As shown in \cref{tab:ablation_heads}, the average performance of the proposed method decreases by 1.0\% and 0.8\% in AUROC and AUPRC respectively when the specialized detection heads are replaced. This shows that specialized detection heads for different synthetic anomaly types help the model capture distinct anomalous characteristics.

\paragraph{Sensitivity of Hyper-parameter $\lambda$ in the Objective Function}
To evaluate the sensitivity of our model to the regularization weight $\lambda$ in \eqref{eq:main_loss}, we conduct experiments across a range of values from 0 to 300. The results are presented with a logarithmic scale for better visualization of performance impacts across different magnitudes of $\lambda$. 

\begin{figure*}[t]
\centering
\includegraphics[width=0.95\linewidth]{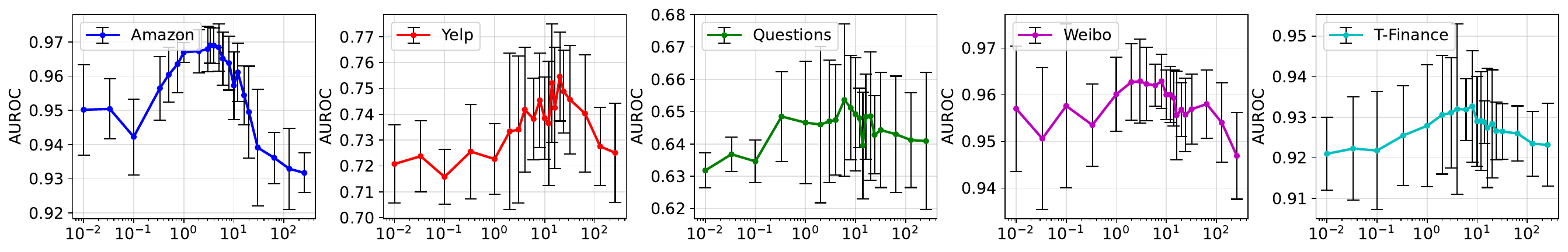}
\caption{Experimental results for investigating the sensitivity of the hyper-parameter $\lambda$.}
\label{fig:regularization_weight_vs_auroc}
\end{figure*}

As shown in \cref{fig:regularization_weight_vs_auroc}, the proposed model tends to perform better when $\lambda$ is in a specific range.
Setting $\lambda$ with a value too high or too low will hinder the model's performance.
As shown in the figure, when $\lambda \in [1, 10]$, the performance remains consistently stable across all datasets, which demonstrates that the model exhibits stable performance within a reasonable range of $\lambda$ values. In the experiments, the value of $\lambda$ is set to 20 for Yelp and 4 for other datasets. The above results show that our hyper-parameter chosen is appropriate in the experiments.

\subsection{Visualization of Synthetic and Real Anomaly Patterns}

\begin{figure}[t]
\centering
\includegraphics[width=0.85\linewidth]{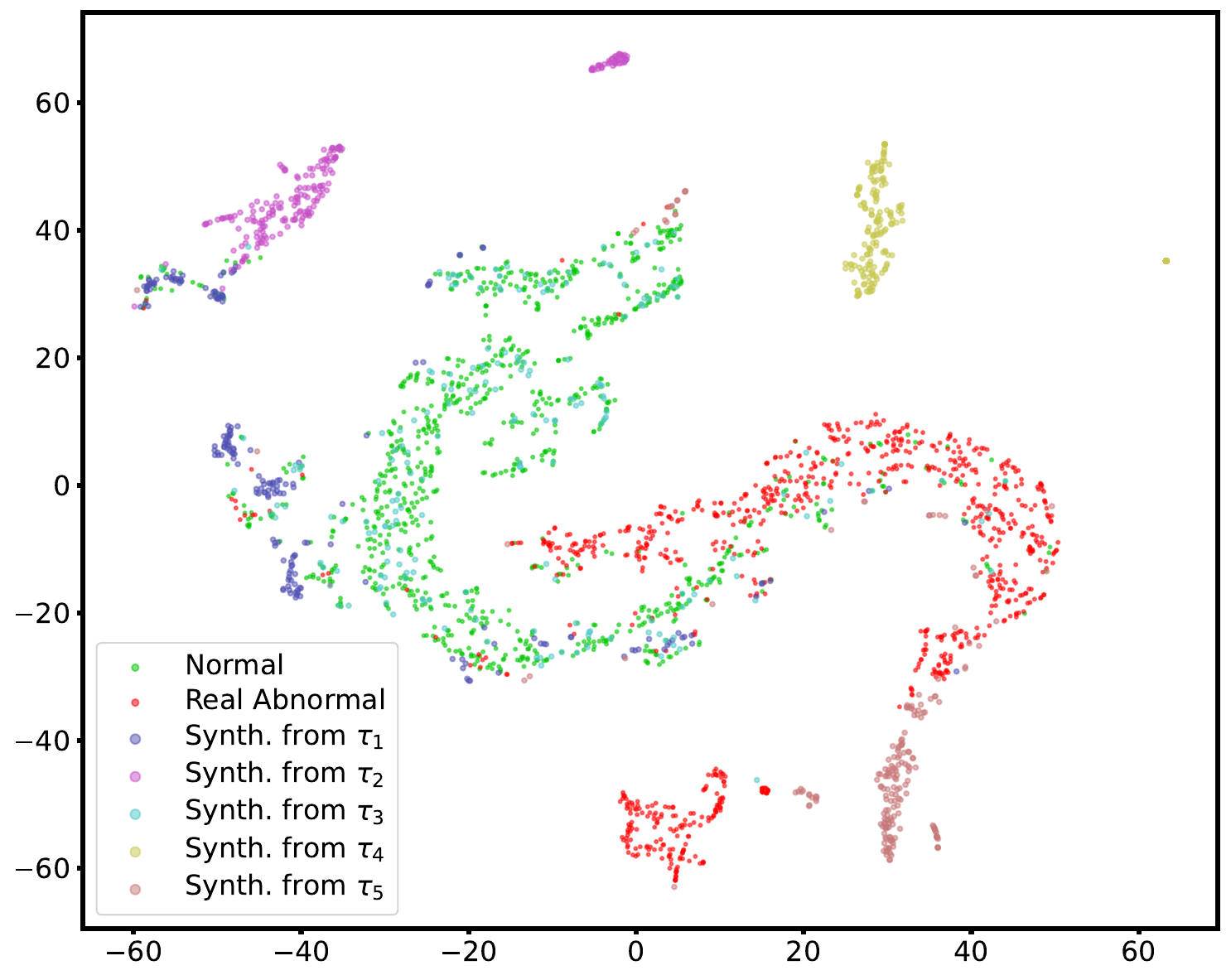}
\caption{t-SNE visualization of learned feature representations for normal nodes, real anomalies, and five types of synthetic anomalies.}
\label{fig:tsne_synthetic_real}
\end{figure}

To visualize the differences between synthetic anomaly patterns and real anomaly patterns, we conducted additional experiments using t-SNE.
As shown in \cref{fig:tsne_synthetic_real}, we visualize the learned feature representations for synthetic and real anomalies, including normal nodes, real anomalies, and all five types of synthetic anomalies from the warm-up phase.
A key observation is that the five synthetic anomaly types and real anomalies occupy different regions in the feature space.
This spatial separation demonstrates that our method does not aim to mimic or replicate real anomaly patterns.
Instead, the model learns to develop sensitivity to perturbations, which enables it to effectively detect real anomalies despite their distinct distributional characteristics from synthetic ones.

\subsection{Comparison of Computational Cost}

To evaluate the computational efficiency of our method, we compare the training and inference time with competing methods on the Amazon dataset.
As shown in \cref{tab:time_complexity}, our two-phase training strategy does incur additional computational overhead during the training phase.
The warm-up phase requires 80.73 seconds, and the full training phase requires 33.16 seconds, resulting in a total training time of 113.89 seconds.
However, compared to some specialized GAD methods like H2-FDetector (1102.00 seconds) and CARE-GNN (134.00 seconds), our total training time remains reasonable while achieving superior detection performance.

Notably, the training cost of the warm-up phase (80.73 seconds) is acceptable when considering the significant performance improvements it brings.
This warm-up overhead is a one-time cost that enables the model to learn disturbance-sensitive features, which substantially enhances detection accuracy across all datasets as demonstrated in our ablation studies. Also, it is important to note that the inference time of our method (0.015 seconds per sample) is comparable to other competing methods including the baseline.
This indicates that once the model is trained, our method does not introduce additional computational burden during deployment, making it practical for real-world applications where inference efficiency is critical.

\begin{table}[t]
\centering
\caption{Training and Inference Time Comparison on Amazon Dataset}
\label{tab:time_complexity}
\renewcommand{\arraystretch}{1.2}
\setlength{\tabcolsep}{3pt}
\small
\begin{tabular}{lccc}
\toprule
\textbf{Model} & \begin{tabular}{c} \textbf{Warm-up} \\ \textbf{(seconds)} \end{tabular} & \begin{tabular}{c} \textbf{Full Training} \\ \textbf{(seconds)} \end{tabular} & \begin{tabular}{c} \textbf{Inference} \\ \textbf{(seconds/sample)} \end{tabular} \\
\midrule
MLP & - & 0.24 & 0.004 \\
GCN & - & 1.21 & 0.010 \\
GraphSAGE & - & 1.06 & 0.011 \\
GAT & - & 3.34 & 0.011 \\
GIN & - & 1.44 & 0.008 \\
\midrule
PC-GNN & - & 0.61 & 0.007 \\
H2-FDetector & - & 1102.00 & 0.097 \\
BWGNN & - & 2.57 & 0.015 \\
GHRN & - & 3.15 & 0.015 \\
CARE-GNN & - & 134.00 & 0.009 \\
GATSep & - & 32.86 & 0.015 \\
CGNN & - & 4.17 & 0.012 \\
\midrule
\textbf{Ours} & 80.73 & 33.16 & 0.015 \\
\bottomrule
\end{tabular}
\end{table}

\section{Conclusion}\label{sec:conclusion}

In this paper, we introduce a weakly-supervised graph anomaly detection method by leveraging a novel encoding strategy that develops disturbance-sensitive feature representation through synthetic anomaly generation and multi-task learning. 
Our method employs a multi-task learning scheme designed to extract robust latent representations, through disturbance-sensitive feature learning based on generated synthetic anomalies.
A two-phase strategy is also designed to train our model. This strategy includes an initial warm-up phase using only synthetic samples, followed by a full-training phase that integrates both synthetic anomalies and real data, which further enhances the performance of graph anomaly detection. 
Through extensive experiments on real-world datasets, we demonstrate that the proposed method significantly outperforms state-of-the-art GAD methods, particularly in scenarios with few labeled anomalies.
The proposed method is primarily designed for static graphs, and does not have the ability to handle anomalies in dynamic graph evolution. Future work could explore the invariant feature representation involving dynamic graph evolution and develop adaptive hyperparameter selection strategies to further enhance the practical applicability of the method.

\section*{Acknowledgments}
The work of Yingjie Zhou, Yuqin Xie and Fanxing Liu was supported in part by National Natural Science Foundation of China (NSFC) under Grant No. 62171302, the 111 Project under Grant No.B21044 and Sichuan Science and Technology Program under Grant No. 2023NSFSC1965.


\bibliographystyle{IEEEtran}
\bibliography{referenceIEEE}


 
\vspace{11pt}

\begin{IEEEbiography}[{\includegraphics[width=1in,height=1.25in,clip,keepaspectratio]{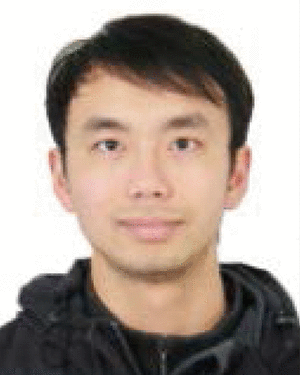}}]{Yingjie Zhou}
(Member, IEEE) received the Ph.D. degree from the School of Communication and Information Engineering, University of Electronic Science and Technology of China (UESTC), China, in 2013. He is currently an Associate Professor with the College of Computer Science, Sichuan University (SCU), China. He was a Visiting Scholar with the Department of Electrical Engineering, Columbia University, New York. His current research interests include behavioral data analysis, network management, and resource allocation. He received the Best Paper Awards at IEEE HPCC and IEEE MMSP. He has served as the Area Chair for IEEE ICASSP 2025, the Program Vice-Chair for IEEE HPCC 2022, the Local Arrangement Chair for IEEE BMSB 2021, etc. He has co-organized AI for Time Series (AI4TS) workshop at IJCAI 2023, 2024. He is an Associate Editor for IEEE Transactions on Emerging Topics in Computational Intelligence.
\end{IEEEbiography}

\vspace{11pt}

\begin{IEEEbiography}[{\includegraphics[width=1in,height=1.25in,clip,keepaspectratio]{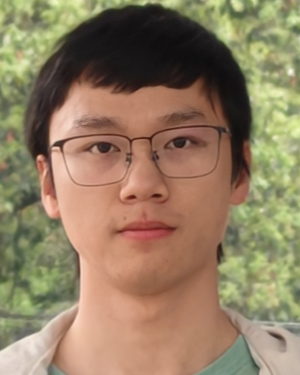}}]{Yuqin Xie}
received the B.S. degree in Software Engineering from Sichuan University, Chengdu, China, in 2023. He is working toward the M.S. degree in Computer Science with the College of Computer Science, Sichuan University, Chengdu, China. His main research interests include data mining and anomaly detection on graph data.
\end{IEEEbiography}

\vspace{11pt}

\begin{IEEEbiography}[{\includegraphics[width=1in,height=1.25in,clip,keepaspectratio]{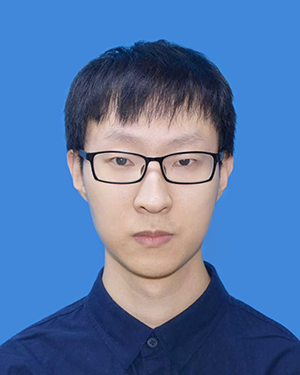}}]{Fanxing Liu}
received the B.S. degree in Software Engineering in 2021 and the M.S. degree in Computer Science and Technology in 2024, both from Sichuan University. He is currently pursuing the Ph.D. degree in Computer Science and Technology at Sichuan University. His main research interests include anomaly detection and behavior analysis.
\end{IEEEbiography}

\vspace{11pt}

\begin{IEEEbiography}[{\includegraphics[width=1in,height=1.25in,clip,keepaspectratio]{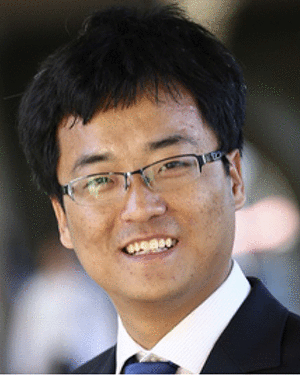}}]{Dongjin Song}
(Member, IEEE) received the Ph.D. degree from the University of California San Diego (UCSD), in 2016. Currently, he is an assistant professor with the School of Computing, University of Connecticut (UConn). His research interests include machine learning, deep learning, data mining, and related applications for time series data and graph representation learning. Papers describing his research have been published at top-tier data science and artificial intelligence conferences, such as NeurIPS, ICML, ICLR, KDD, ICDM, SDM, AAAI, IJCAI, CVPR, ICCV, etc. He has co-organized AI for Time Series (AI4TS) workshop at IJCAI 2022, 2023 and the Mining and Learning from Time Series workshop at KDD 2022, 2023. He has also served as senior PC for AAAI, IJCAI, and CIKM. He won the UConn Research Excellence Research (REP) Award in 2021. His research has been funded by NSF, USDA, Morgan Stanley, NEC Labs America, Travelers, etc.
\end{IEEEbiography}

\vspace{11pt}

\begin{IEEEbiography}[{\includegraphics[width=1in,height=1.25in,clip,keepaspectratio]{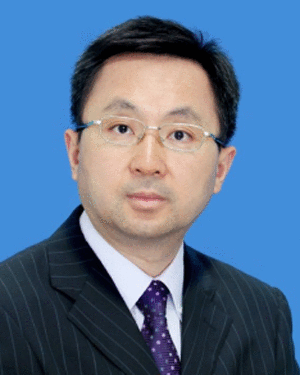}}]{Ce Zhu}
(Fellow, IEEE) received the B.S. degree in electronic and information engineering from Sichuan University, Chengdu, China, in 1989, and the M.Eng. and Ph.D. degrees in electronic and information engineering from Southeast University, Nanjing, China, in 1992 and 1994, respectively. He was a Postdoctoral Researcher with The Chinese University of Hong Kong, Hong Kong, in 1995; the City University of Hong Kong, Hong Kong; and The University of Melbourne, Melbourne, VIC, Australia, from 1996 to 1998. He was with Nanyang Technological University, Singapore, from 1998 to 2012, for 14 years, where he was a Research Fellow, a Program Manager, an Assistant Professor, and then promoted to an Associate Professor in 2005. Since 2012, he has been with the University of Electronic Science and Technology of China, Chengdu, as a Professor. His research interests include video coding and communications, video analysis and processing, 3D video, and visual perception and applications. 

Dr. Zhu was a co-recipient of multiple paper awards at international conferences, including most recently the Best Demo Award in IEEE MMSP 2022 and the Best Paper Runner-Up Award in IEEE ICME 2020. He has served on the editorial boards for a few journals, including an Associate Editor for IEEE Transactions on Image Processing, IEEE Transactions on Circuits and Systems for Video Technology, IEEE Transactions on Broadcasting, and IEEE Signal Processing Letters; an Editor for IEEE Communications Surveys and Tutorials; and an Area Editor for Signal Processing: Image Communication. He has also served as a Guest Editor for a few special issues in international journals, including a Guest Editor for IEEE Journal of Selected Topics in Signal Processing. He was an APSIPA Distinguished Lecturer, from 2021 to 2022, and an IEEE Distinguished Lecturer of Circuits and Systems Society, from 2019 to 2020.
\end{IEEEbiography}

\vspace{11pt}

\begin{IEEEbiography}[{\includegraphics[width=1in,height=1.25in,clip,keepaspectratio]{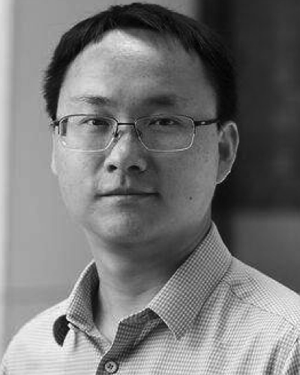}}]{Lingqiao Liu}
(Member, IEEE) received the B.S. and M.S. degrees in communication engineering from the University of Electronic Science and Technology of China, Chengdu, in 2006 and 2009, respectively, and the Ph.D. degree from the Australian National University, Canberra, in 2014. He is currently a Senior Lecturer with The University of Adelaide and the Australian Institute for Machine Learning. His current research interests include low-supervision learning and various topics in computer vision and natural language processing. In 2016, he was awarded the Discovery Early Career Researcher Award from the Australian Research Council and the University Research Fellow from The University of Adelaide.
\end{IEEEbiography}

\vspace{11pt}

\vfill

\end{document}